\documentclass[conference]{IEEEtran}
\IEEEoverridecommandlockouts

\usepackage[T1]{fontenc}
\usepackage{cite}
\usepackage{amsmath,amssymb,amsfonts}
\usepackage{algorithmic}
\usepackage{graphicx}
\usepackage[final]{changes}

\usepackage{xcolor}
\usepackage{booktabs}
\usepackage{url}
\def\BibTeX{{\rm B\kern-.05em{\sc i\kern-.025em b}\kern-.08em
    T\kern-.1667em\lower.7ex\hbox{E}\kern-.125emX}}

\makeatletter
\newcommand{\linebreakand}{%
  \end{@IEEEauthorhalign}
  \hfill\mbox{}\par
  \mbox{}\hfill\begin{@IEEEauthorhalign}
}

\begin{document}

\title{LoRA Enhanced Contrastive Learning with SAS Vision Transformers
\thanks{This work was supported by the Office of Naval Research (N0001426GI00772) and their Internal Applied Research  program.}
}

\author{\IEEEauthorblockN{Dan Zimmerman}
\IEEEauthorblockA{\textit{Center for Connected Autonomy \& AI} \\
\textit{Florida Atlantic University}\\
Boca Raton, USA \\
dzimmerman2021@fau.edu}
\and
\IEEEauthorblockN{Frank E. Bobe III}
\IEEEauthorblockA{\textit{Naval Surface Warfare Center} \\
\textit{Panama City Division}\\
Panama City, USA \\
frank.e.bobe.civ@us.navy.mil}
\and
\IEEEauthorblockN{Amelia L. McCormack}
\IEEEauthorblockA{\textit{Department of Computer Science} \\
\textit{Florida State University}\\
Tallahassee, USA \\
alm20h@fsu.edu}
\linebreakand
\IEEEauthorblockN{Matthew Cook}
\IEEEauthorblockA{\textit{Naval Surface Warfare Center} \\
\textit{Panama City Division}\\
Panama City, USA \\
matthew.g.cook12.civ@us.navy.mil}
\and
\IEEEauthorblockN{Gregory D. Vetaw}
\IEEEauthorblockA{\textit{Naval Surface Warfare Center} \\
\textit{Panama City Division}\\
Panama City, USA \\
gregory.d.vetaw.civ@us.navy.mil}

}

\maketitle

\begin{abstract}
Automatic target recognition (ATR) with synthetic aperture sonar (SAS) enables
advanced naval capabilities, but deep learning approaches remain constrained by
human-in-the-loop assessment and by the extreme scarcity of imaged targets
against background clutter. We present a three-stage parameter-efficient
adaptation framework with contrastive refinement, adapting self-DIstillation
with NO labels v3 (DINOv3) Vision Transformer (ViT) models to underwater SAS
ATR. Stage 1 applies a Low-Rank Adaptation (LoRA) workflow to bridge the gap
between pretraining on natural images and the physics of underwater acoustic
propagation while freezing the ViT backbone. Stage 2 implements a ``Refiner''
using hard negative mining to harden the decision boundary against acoustic
mimics: naturally occurring seafloor features, such as rock outcrops and
sediment formations, whose sonar signatures resemble man-made targets. Stage 3
applies Supervised Contrastive Learning (SupCon) to pull targets into a compact
manifold away from clutter. We evaluate on at-sea SAS data under a
mission-level geographic split, read every arm at a matched $85\%$ test recall,
and repeat each comparison over three random seeds. Adaptation accounts for the
entire effect: LoRA raises the Area Under the Precision-Recall Curve (AUPRC)
from $0.300$ to $0.679 \pm 0.027$ over the identical frozen backbone, a $+0.38$
gap an order of magnitude beyond any other effect we measure, and rank $4$
reaches the same result while training only $0.26\%$ of its weights. Neither
refinement stage separates from its matched control: mining is worth
$-0.0045 \pm 0.0119$ AUPRC against a random curriculum of equal size, and
SupCon $+0.0002 \pm 0.0096$ against the stage it refines. We report both null
results with their mechanisms, namely that the curriculum is mined on a split
the encoder has already fit, and that the supervised stages impose most of the
target--clutter geometry before the contrastive term is applied. One efficient
adaptation stage is sufficient; the curriculum stacked on top of it is not.
\end{abstract}

\begin{figure*}[t]
    \centering
    \includegraphics[width=0.62\textwidth]{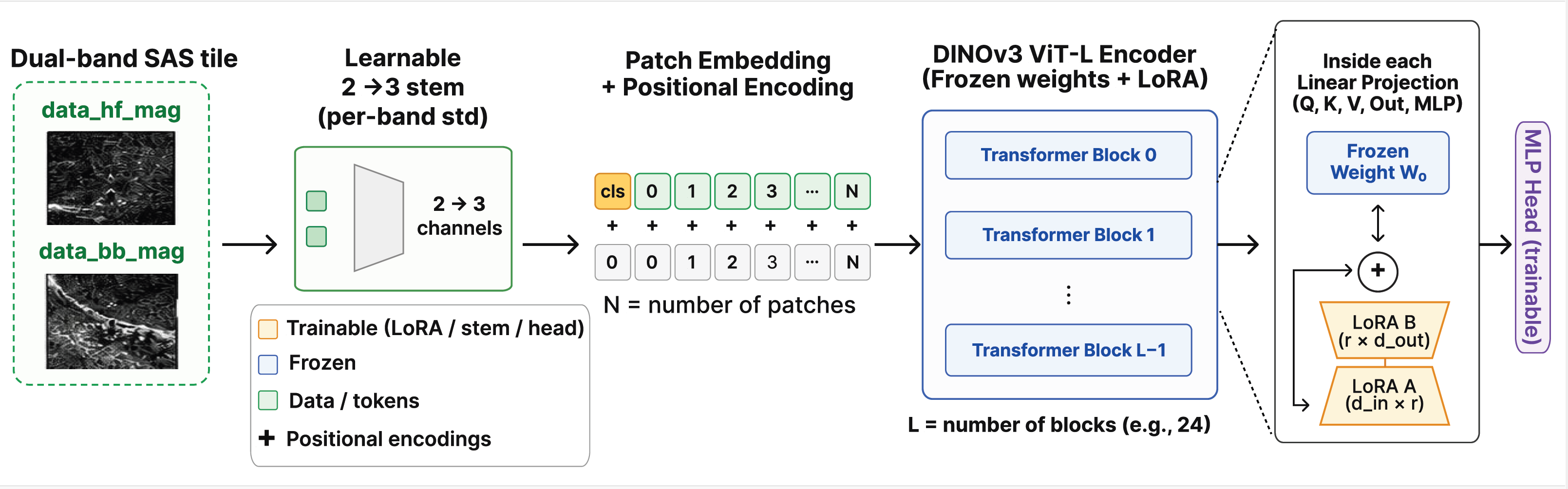}
    \caption{The parameter-efficient fine-tuning (PEFT) architecture used for adapting the DINOv3 ViT-L model. Dual-band (\replaced{High-Frequency (HF) and Broadband (BB)}{HF/BB}) SAS imagery is passed through a trainable 2-to-3 channel stem. The frozen ViT\added{-L} backbone is augmented with trainable LoRA modules in its linear projection layers.}
    \label{fig:pipeline_wide}
\end{figure*}

\section{Introduction}
\label{sec:introduction}
Automatic target recognition (ATR) enables advanced naval capabilities for underwater and maritime applications~\cite{Sledge_circular_detection, Vetaw, Gerg_autofocus, Yang2026FS2DETRTF}, including identification of objects that endanger littoral operations~\cite{Kohntopp_classification, Williams_spectral, Williams_classification}. Supervised deep learning has advanced synthetic aperture sonar (SAS) analysis, but it relies on massive human-verified datasets, and in operational environments target imagery is scarce against a vast background of clutter. Much of that clutter consists of \emph{acoustic mimics}: naturally occurring seafloor features, such as rock outcrops, boulders and sediment formations, whose highlight-and-shadow signatures resemble those of man-made targets. These features degrade standard classifiers and compound the underlying class imbalance~\cite{RCDIYOLO2025}.

To bypass this bottleneck we introduce a parameter-efficient domain adaptation framework for SAS ATR. The framework leverages a DINOv3~\cite{DINOv3_2025} Vision Transformer (ViT-L) foundation model pretrained on natural imagery, which does not capture the acoustic propagation physics of sonar. To bridge the gap between natural-image pretraining and the physics underlying underwater acoustic propagation, we propose a three-stage Low-Rank Adaptation (LoRA)~\cite{LoRA_2021} workflow comprising: (1) an acoustic domain adaptation stage that aligns the ViT-L with foundational sonar features; (2) an active-learning ``Refiner'' that uses hard negative mining to map the decision boundaries of complex seafloor textures; and (3) a Supervised Contrastive Learning (SupCon)~\cite{khosla2021} stage that projects the data into a separable geometric feature space.

Crucially, we evaluate each stage against a \emph{matched control}: a run identical in curriculum size, optimizer, schedule and weight initialization, differing only in the single mechanism that stage contributes. Comparing a stage against its own control, rather than against the final end-to-end accuracy of the assembled pipeline, isolates the effect of the mechanism from the effect of longer training. Every arm is read at a matched 85\% test recall, and we repeat every close comparison across three random seeds.

This protocol changes the conclusion. Adapting the frozen backbone with LoRA raises the Area Under the Precision-Recall Curve (AUPRC) by more than a factor of two while training at most $4.2\%$ of the backbone's weights, and as little as $0.26\%$ suffices; neither subsequent stage separates from its control. We report those two null results in full rather than omitting them, because each is accompanied by a mechanism that explains it: the mined curriculum is selected on a split the encoder has already fit, and the supervised stages impose most of the target--clutter geometry before the contrastive term is applied. The practical consequence is that engineering effort belongs in the single adaptation step rather than in the multi-stage curriculum built on top of it.

Our contributions are: (i) a parameter-efficient adaptation of DINOv3 to dual-band SAS recovering a $+0.38$ AUPRC gap over the same frozen backbone; (ii) evidence that adapter capacity saturates at very low rank and that low-rank adaptation is not separable from full fine-tuning at either backbone scale we tested; and (iii) a controlled, matched-control accounting of two widely assumed pipeline components, hard negative mining and supervised contrastive refinement, establishing that neither repays its cost at this data scale and supplying the mechanism behind each null result.

\section{Related Work}
\label{sec:related_work}
\textbf{Deep learning for SAS ATR.} SAS ATR has relied on supervised convolutional neural networks (CNNs)~\cite{ResNet_2016} to extract spatial features from acoustic imagery~\cite{Williams_classification, Sledge_circular_detection}, extended to spectral partitioning~\cite{Williams_spectral}, volumetric normalization~\cite{Vetaw}, and seafloor classification~\cite{Kohntopp_classification}. Lightweight detectors, including optimized You Only Look Once (YOLO) variants~\cite{RCDIYOLO2025, SS-YOLO_2025} and few-shot transformer detectors~\cite{Yang2026FS2DETRTF}, have advanced real-time analysis. The extreme imbalance between targets and clutter nonetheless remains a persistent challenge. 

\textbf{Foundation models and parameter-efficient adaptation.} Vision Transformers~\cite{ViT_2020} pretrained with the DINOv2~\cite{DINOv2_2023} and DINOv3~\cite{DINOv3_2025} self-supervised frameworks
generalize strongly across optical imagery~\cite{ILSVRC15}, but the shift to underwater acoustic propagation is challenging. Full fine-tuning at this size is prohibitive without distributed optimizations~\cite{DeepSpeed_2020, ZeRO_2020}. Parameter-efficient fine-tuning~\cite{PEFT_2022} has emerged as an effective strategy for adapting large ViTs to specialized domains such as SAS ATR. LoRA~\cite{LoRA_2021} freezes the pretrained weights and injects trainable low-rank updates that substantially reduce the computational budget. Later variants improve its flexibility by training nested ranks~\cite{NoRA_2024} or adjusting the rank during training~\cite{DyLoRA_2023} rather than fixing it in advance. 

\textbf{Contrastive learning and extreme class imbalance.} Re-sampling~\cite{SMOTE_2002} and cost-sensitive objectives~\cite{FocalLoss_2017} are limited at sonar imbalance ratios. Contrastive frameworks~\cite{SimCLR_2020, ContrastiveClustering_2020} instead maximize agreement between augmented views, and SupCon~\cite{khosla2021} extends this with labels. Pairing SupCon with hard-negative mining~\cite{HardNeg_2024} can enforce separability that cross-entropy struggles to achieve. 
This work builds on an earlier investigation of ViT-L backbones and contrastive loss for SAS~\cite{Team_SAS_ViT_Unpub}, and reports the Matthews correlation coefficient (MCC)~\cite{MCC_1975} as its imbalanced-data metric.

\section{Methodology}

We adapt the DINOv3 ViT for SAS ATR in a \textbf{three-stage pipeline}, moving from broad feature alignment to adversarial refinement and finally to geometric contrastive regularization.

\subsection{Stage 1: Foundational LoRA Domain Adaptation}
To bridge the domain gap between \deleted{the optical pretraining of the frozen DINOv3 ViT-L backbone} \added{pretraining a DINOv3 ViT-L backbone on massive amounts of optical imagery} and dual-band acoustic physics, we implement the parameter-efficient fine-tuning (PEFT) architecture of Fig.~\ref{fig:pipeline_wide}. Because the backbone expects three channels, a trainable $1\times1$ convolutional 2-to-3 channel ``stem'' projects the dual-band high frequency (HF) and broadband (BB) SAS magnitude channels into a compatible space. Trainable Low-Rank Adaptation (LoRA) modules are then injected into the query ($Q$), key ($K$), value ($V$), and output ($O$) linear projections of each self-attention block. The placement is \emph{attention-only}; that is, the feed-forward projections carry no adapter, so every result here is obtained by re-weighting attention alone. 

This architecture is trained in two ways that differ only in the objective and in the role of the \replaced{Multilayer Perceptron (MLP)}{MLP}. \textbf{The supervised form is the pipeline's Stage 1 throughout this paper: it is what Stages 2 and 3 build upon and what appears in the Stage 1 rows of Table~\ref{tab:performance_comparison}.} There, the MLP is a binary classifier over the final \replaced{Classify token (CLS)}{CLS} embedding, trained with the Asymmetric Loss of Ridnik et al.~\cite{ASL_2021} over class-balanced batches, so labels enter both the objective and the sampler, and Stage 2 initializes from this encoder.

The self-supervised form is a separate comparison \replaced{module}{arm},
\added{which is} never \added{used as an} \deleted{the} input to a later stage. The same MLP acts instead as a projection head onto a 128-dimensional unit hypersphere optimized with the Contrastive Clustering objective~\cite{ContrastiveClustering_2020} over two augmented views. That objective is not purely instance-level. It combines an instance \replaced{Information Noise-Contrastive Estimation (InfoNCE)}{InfoNCE} term, as in \replaced{Simple Framework for Contrastive Learning of Visual Representations (SimCLR)}{SimCLR}, with a cluster-assignment head ($K = 50$) grouping the batch without labels, which are used only for sampling. Its frozen features are scored by a linear probe, appearing as the \replaced{Contrastive Clustering Self-Supervised Learning (CC-SSL)}{CC-SSL} row of Table~\ref{tab:performance_comparison}.

\subsection{Stage 2: Adversarial Refinement}
Performance after Stage 1 is limited by geometrically complex clutter. To harden the decision boundary against these ``acoustic mimics,'' \deleted{(Fig.~\ref{fig:stages_wide})} we freeze the Stage 1 encoder, extract embeddings for the training set, train a lightweight linear probe on those frozen features to score each sample, and take the top $K = 5{,}000$ clutter samples by false-positive confidence as hard negatives. We then build a curriculum of all known target signatures plus those negatives, initializing from the Stage 1 adapters and band stem rather than at random so the network continues adapting its attention subspace, and fine-tune at $2 \times 10^{-5}$ with the same Asymmetric Loss~\cite{ASL_2021}. Because this curriculum alters the class balance regardless of \emph{which} clutter is chosen, mining confounds the informativeness of the selected negatives with simple exposure rebalancing. We therefore pair every Stage 2 run with a \textbf{random exposure control}, identical but for drawing its $5{,}000$ snippets uniformly rather than by score, so the difference between the arms isolates the mining signal.

\subsection{Stage 3: Contrastive Geometric Regularization}
To resolve the remaining ambiguity between targets and acoustic mimics, we introduce a Supervised Contrastive Learning (SupCon) stage~\cite{khosla2021}, treating the embedding space as a geometric problem that structurally enforces class separability. \deleted{(Fig.~\ref{fig:stages_wide})} Initializing from the refined Stage 2 adapters, stem and classifier head, we continue training on the hard-negative curriculum with the dual-loss objective $\mathcal{L}_{\text{total}} = \lambda_{\text{sup}}\mathcal{L}_{\text{sup}} + \lambda_{\text{cls}}\mathcal{L}_{\text{cls}}$, where $\mathcal{L}_{\text{cls}}$ is the imbalance-aware asymmetric loss over the classification logits of two stochastically augmented views per sample, and $\mathcal{L}_{\text{sup}}$ is the Supervised Contrastive loss~\cite{khosla2021}. A temporary projection head maps representations to a 128-dimensional unit hypersphere, and for a batch of $N$ samples stochastically doubled to a multiview batch of $2N$ views,
\begin{equation}
    \mathcal{L}_{\text{sup}} = \sum_{i \in I} \frac{-1}{|P(i)|} \sum_{p \in P(i)} \log \frac{\exp(z_i \cdot z_p / \tau)}{\sum_{a \in A(i)} \exp(z_i \cdot z_a / \tau)},
\end{equation}
with $I \equiv \{1, \dots, 2N\}$ indexing the multiview batch, $z_i$ the projected anchor, $P(i)$ the positive views sharing its class label, $A(i) \equiv I \setminus \{i\}$ is the index set of all other views in the batch, and $\tau$ is the
temperature parameter (set to $0.07$). Crucially for the analysis that follows, $\mathcal{L}_{\text{sup}}$ is not minimized at zero: for a perfectly class-collapsed embedding it attains $\mathbb{E}_i[\log|P(i)|]$, a constant fixed by batch composition, so its distance above that floor, not its raw value, indicates how much geometric structure the objective still has to impose.

\section{Implementation and Training Details}
\label{sec:training_details}
\subsection{Dataset and Splits}
We evaluate on dual-band (High-Frequency and Broadband) SAS magnitude imagery from at-sea deployments, comprising 148 mission files partitioned at the \emph{mission} level by geographic location, so no mission contributes to more than one split and the test set represents a genuine geographic shift rather than a random sample of the same surveys. Training comprises 106 missions and 336,298 snippets containing 2,791 targets, which equates to a severe $119{:}1$ imbalance; validation and test contain 16 and 26 missions, at 36,327 snippets ($99{:}1$) and 67,083 snippets ($155{:}1$).

\subsection{Preprocessing and Augmentation}
Imagery is bilinearly resized to $224 \times 224$ to match the DINOv3 ViT-L input resolution and phase is discarded. Each band is standardized independently to zero mean and unit variance, clamped at $\pm 8$ standard deviations to tame the heavy tail from bright target returns, identically at every stage. For the contrastive phases we apply Rayleigh-distributed speckle noise, random resized cropping, flipping, photometric jitter and Gaussian blurring ($\sigma \in [0.1, 2.0]$), so the loss is driven by target geometry rather than pixel similarity.

\subsection{Optimization and Model Selection}
The pipeline is implemented in PyTorch and trained using distributed data-parallel processing across eight NVIDIA A6000 GPUs, giving a pooled batch of 1,024. DINOv3 ViT-L weights remain frozen throughout. LoRA modules use rank $r=64$ with $\alpha = 2r$, selected by a sweep over $r \in \{2, 4, 8, 16, 32, 64\}$ on validation AUPRC.

\textbf{Stage 1} uses Adam at $1 \times 10^{-4}$, batch sizes of 128 per GPU, and the Asymmetric Loss~\cite{ASL_2021} over class-balanced batches; the self-supervised comparison variant substitutes a Contrastive Clustering loss with instance temperature $\tau_i = 0.5$ and a $K = 50$ cluster head. \textbf{Stage 2} fine-tunes the 5,000-hard-negative curriculum with AdamW at $2 \times 10^{-5}$ over 30 epochs with cosine annealing, again under the Asymmetric Loss; the random-selection control is identical except that its 5,000 clutter snippets are drawn uniformly. \textbf{Stage 3} initializes from Stage 2 and trains dual-view batches with AdamW at $2 \times 10^{-5}$, SupCon temperature $\tau = 0.07$ and a 128-dimensional projection head. Both loss weights are unity, $\lambda_{\text{sup}} = \lambda_{\text{cls}} = 1$. Sweeping $\lambda_{\text{sup}}$ over $\{0, 0.1, 0.3, 3, 10\}$ moves blind-test AUPRC by at most $0.0074$, inside the $\pm 0.0096$ paired seed interval, and disabling the term outright costs $0.0041$.

Model selection at every stage uses validation AUPRC, with early stopping after ten checks without improvement and the best checkpoint restored before a frozen evaluation pass; Section~\ref{sec:limitations} discusses the resulting selection bias. Every comparison smaller than the seed-to-seed spread of roughly $\pm 0.03$ AUPRC was repeated at seeds 42, 100 and 999 and is reported as a mean with a $95\%$ confidence interval, or as a paired per-seed difference where available.

\section{Results and Performance Analysis}
\label{sec:results}
All results are on an unseen test split (67,083 snippets, 431 targets). No label from the validation or test missions reaches any arm at any stage; the one qualification, concerning unlabeled imagery, is in Section~\ref{sec:limitations}. Every arm shares one backbone, one split and one evaluation harness, and all results are means over three seeds (42, 100 and 999) with $95\%$ confidence intervals.

\subsection{Evaluation Protocol}
\label{subsec:protocol}
All threshold-based metrics are read at a \textbf{recall-matched operating point}: the threshold is placed so each model reaches 85\% recall on the blind test set. MCC and $n_{\text{FA}}$ are the threshold-based metrics reported here, and we quote the False Positive Rate at that anchor ($\text{FPR}_{85}$) as the headline false-positive rate, alongside the threshold-free AUPRC. Where a claim concerns the high-recall regime specifically, we report $\text{FPR}_{95}$ and say so.

\begin{table*}[t]
\centering
\caption{Blind-test performance. Mean $\pm$ 95\% CI over three seeds (42, 100, 999) for every arm. MCC$^{*}$ and $n_{\text{FA}}$ are read at a matched 85\% test recall; AUPRC and AUROC are threshold-free. No entry is emboldened: the adapted arms have overlapping intervals and are not separable.}
\label{tab:performance_comparison}
\scriptsize
\setlength{\tabcolsep}{7pt}
\begin{tabular}{@{}lcccc@{}}
\toprule
Model  & AUPRC & AUROC & MCC$^{*}$ & $n_{\text{FA}}$ \\ \midrule
ResNet18, scratch             & 0.563 $\pm$ 0.020 & 0.982 $\pm$ 0.014 & 0.534 $\pm$ 0.065 & 1141 $\pm$ 207 \\
ResNet18, IN-1k              & 0.590 $\pm$ 0.053 & 0.981 $\pm$ 0.007 & 0.570 $\pm$ 0.051 & 841 $\pm$ 218 \\
TinyViT-21M, scratch          & 0.546 $\pm$ 0.039 & 0.980 $\pm$ 0.012 & 0.512 $\pm$ 0.045 & 1539 $\pm$ 271 \\
TinyViT-21M, IN-22k distilled  & 0.640 $\pm$ 0.085 & 0.991 $\pm$ 0.007 & 0.605 $\pm$ 0.064 & 712 $\pm$ 235 \\ \midrule
ViT-L, frozen                 & 0.300 $\pm$ 0.011 & 0.975 $\pm$ 0.001 & 0.350 $\pm$ 0.010 & 3555 $\pm$ 156 \\
ViT-L, full fine-tune         & 0.698 $\pm$ 0.046 & 0.989 $\pm$ 0.008 & 0.640 $\pm$ 0.065 & 615 $\pm$ 304 \\
ViT-L, CC-SSL linear probe    & 0.635 $\pm$ 0.023 & 0.993 $\pm$ 0.000 & 0.549 $\pm$ 0.043 & 897 $\pm$ 253 \\ \midrule
Stage 1, LoRA $r=4$           & 0.686 $\pm$ 0.033 & 0.990 $\pm$ 0.003 & 0.641 $\pm$ 0.039 & 631 $\pm$ 201 \\
Stage 1, LoRA $r=64$          & 0.679 $\pm$ 0.027 & 0.991 $\pm$ 0.001 & 0.626 $\pm$ 0.039 & 643 $\pm$ 83 \\
Stage 2, mined HNM            & 0.682 $\pm$ 0.007 & 0.989 $\pm$ 0.001 & 0.638 $\pm$ 0.002 & 675 $\pm$ 86 \\
Stage 2, random control      & 0.686 $\pm$ 0.017 & 0.990 $\pm$ 0.001 & 0.632 $\pm$ 0.028 & 637 $\pm$ 203 \\
Stage 3, SupCon              & 0.682 $\pm$ 0.016 & 0.986 $\pm$ 0.013 & 0.643 $\pm$ 0.015 & 681 $\pm$ 167 \\ \bottomrule
\end{tabular}
\end{table*}

\begin{figure}[t]
\centering
\includegraphics[width=0.7\columnwidth]{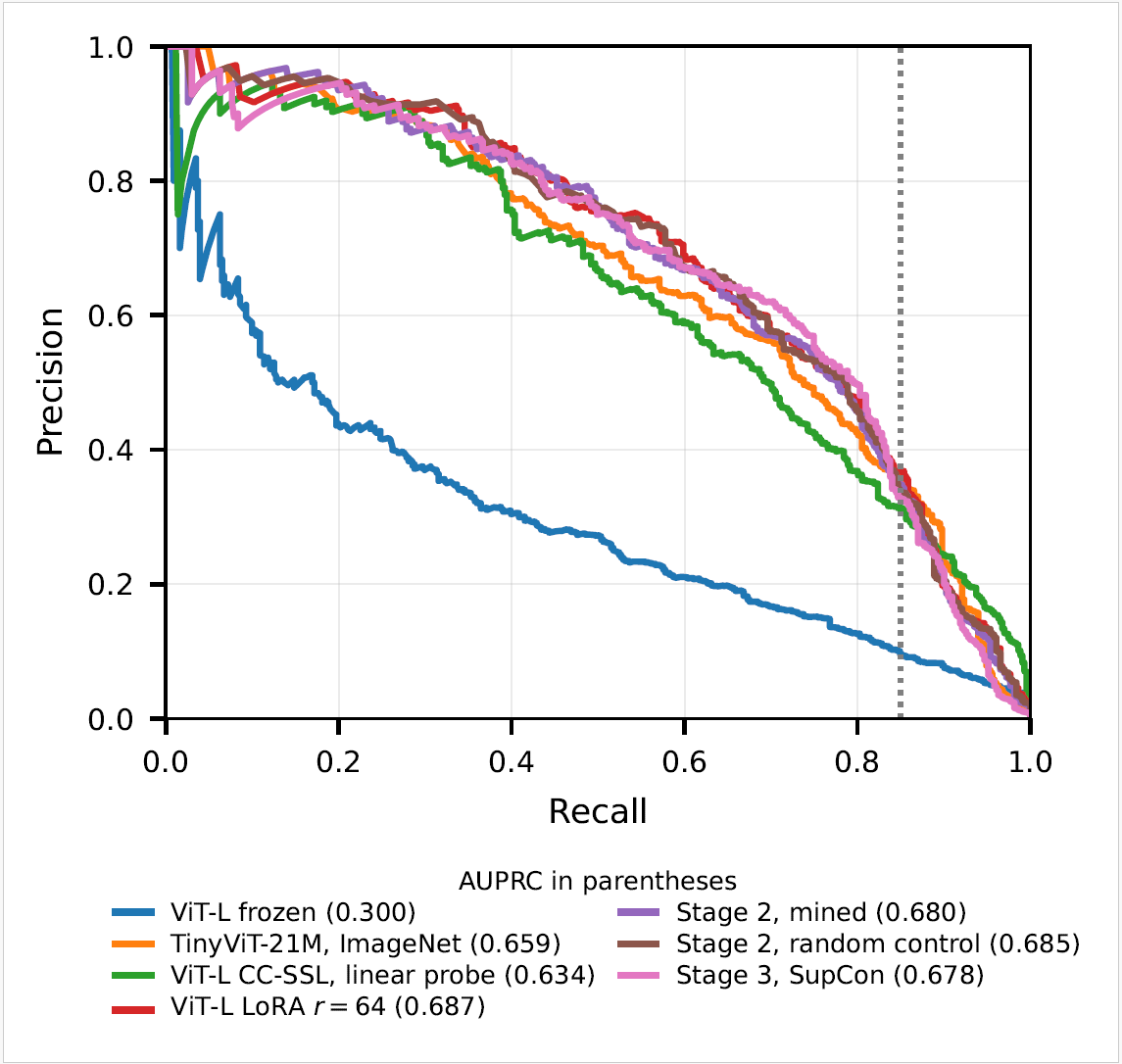}
\caption{Precision-recall on the blind test split, seed 42. The dotted rule marks the $85\%$ recall anchor at which Table~\ref{tab:performance_comparison} is read.}
\label{fig:pr_curves}
\end{figure}

\begin{figure}[t]
\centering
\includegraphics[width=0.7\columnwidth]{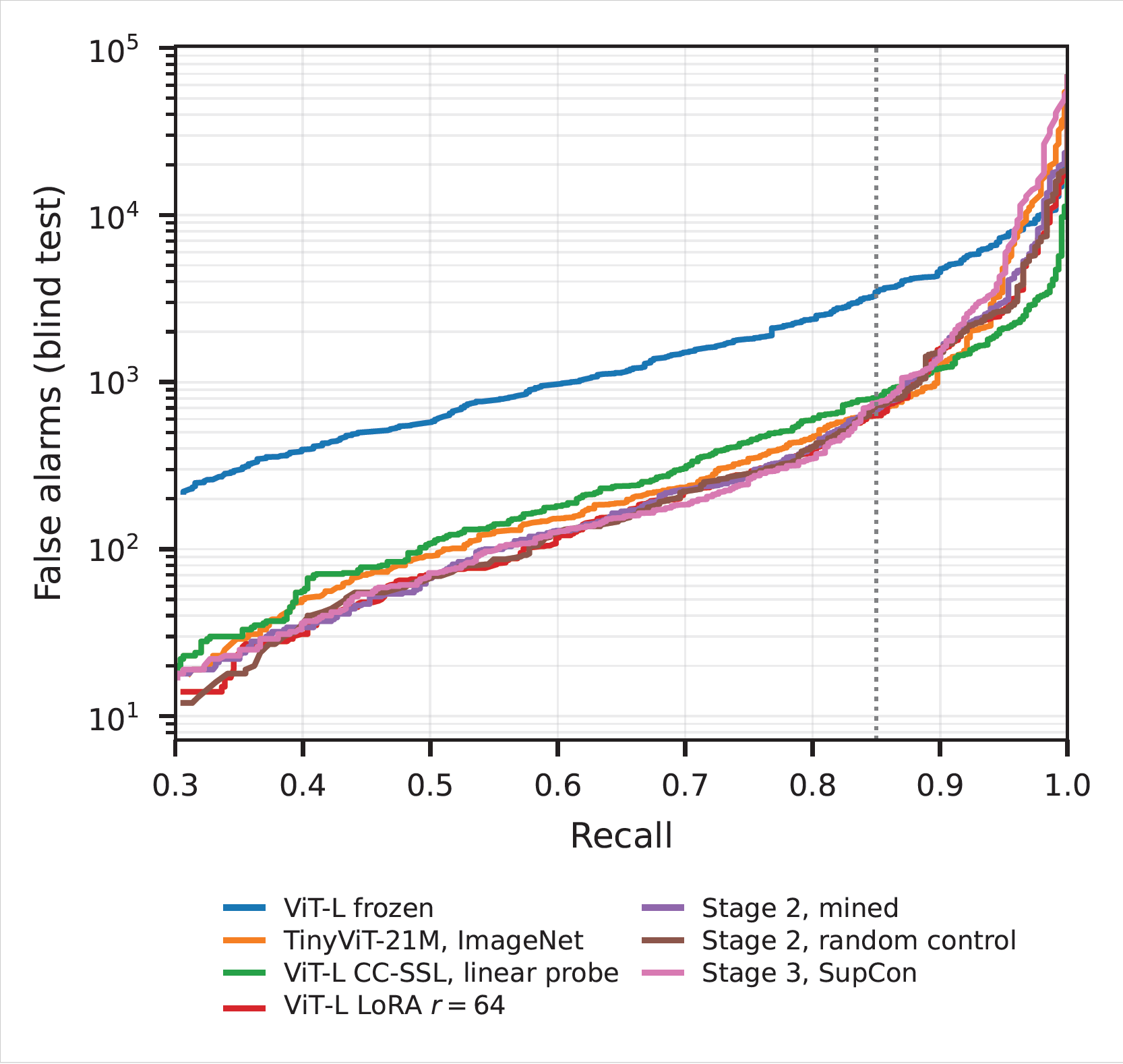}
\caption{False alarms against recall on the blind test split, seed 42, logarithmic ordinate. Derived from the same precision-recall data and verified against the tabulated counts at the operating point.}
\label{fig:nfa_vs_recall}
\end{figure}

\subsection{Domain Adaptation is the Dominant Effect}
\label{subsec:adaptation}
Table~\ref{tab:performance_comparison} isolates each component's contribution, and the largest effect by a wide margin is the domain adaptation itself. The frozen DINOv3 ViT-L, probed linearly over unadapted features, reaches $0.300 \pm 0.011$ AUPRC, worse than a ResNet18 trained from scratch; injecting LoRA adapters into the same frozen backbone raises this to $0.679 \pm 0.027$, a $2.3\times$ improvement. This indicates that optical features do not transfer to acoustic imagery: the adaptation, not the backbone, carries the result.

Comparing against the conventional baselines shows a different picture. We pair results over the same seeds, since multi-seed results are not directly comparable to the single-seed numbers we reported previously. Paired this way, the adapted ViT-L leads the strongest baseline, TinyViT-21M pretrained on ImageNet-22k with distillation, by $+0.046 \pm 0.113$ AUPRC at $r = 4$ and $+0.039 \pm 0.100$ at $r = 64$; both intervals span zero, although the adapted arm is ahead at all three seeds. Pretraining behaves the same way, worth $+0.094 \pm 0.098$ for TinyViT and $+0.027 \pm 0.035$ for ResNet18 against their from-scratch counterparts, again positive at every seed and again not resolvable. We therefore claim only what separates: \textbf{the adapted foundation model is not distinguishable from a well-initialized conventional CNN at this sample size}. The comparison is not like-for-like on pretraining either, with DINOv3 carrying $1.689$B web images against roughly $14$M for TinyViT and $1.28$M for ResNet18. That the frozen ViT-L nonetheless scores $0.300$ is the sharper observation: two orders of magnitude more pretraining data buys nothing until it is adapted.

\subsection{Adapter Capacity Saturates, and Matches Full Fine-Tuning}
\label{subsec:rank}
We swept the LoRA rank over $r \in \{2, 4, 8, 16, 32, 64\}$ and selected $r=64$ on validation AUPRC. On the blind test set performance is \emph{flat} across this range: paired per-seed, $r=4$ against $r=64$ gives $+0.0061 \pm 0.0465$ AUPRC with $r=4$ nominally ahead, and only $r=2$ falls away, to $0.6229$. So $786$K trainable parameters, $0.26\%$ of the backbone, match the $12.58$M of $r=64$, an empirical confirmation of the low intrinsic rank hypothesis of Hu et al.~\cite{LoRA_2021} in the acoustic domain.

An earlier reading of these experiments, drawn from single seeds, held that low-rank adaptation matches full fine-tuning only above some backbone size; replicating every arm at three seeds removes that condition. On the small architectural variant (ViT-S) ($0.612 \pm 0.050$ at $r=64$) a full fine-tune of all 21.6M parameters is worth a paired $-0.001 \pm 0.151$; at the Large scale (ViT-L), fine-tuning all 300M gives $0.698 \pm 0.046$ against $0.679 \pm 0.027$, a paired $+0.018 \pm 0.060$. Neither scale separates, and the rank curve is flat at both. The apparent ViT-S steepening we previously reported came from a single seed whose $r=4$ draw was the lowest of its three.

The LoRA-adapted ViT-L is stable across experiments: \textbf{adapting $0.26\%$ of a ViT-L, or $0.7\%$ of a ViT-S, is not distinguishable from updating every weight}, and rank beyond $r{=}4$ buys nothing at either size. One asymmetry emerges: full fine-tuning is the noisier regime, with a seed standard deviation of $0.019$ against $0.011$ at ViT-L and $0.042$ against $0.020$ at ViT-S, so equal expected performance at roughly half the variance is the practical case for the adapter.

\subsection{Hard Negative Mining Does Not Separate From Random Selection}
\label{subsec:hnm}
Stage 2 mines the $K=5{,}000$ clutter snippets that the Stage 1 encoder scores as most target-like and fine-tunes on those plus all known targets; to test whether mining contributes anything beyond rebalancing exposure, we ran an otherwise identical control drawing its $5{,}000$ snippets \emph{at random}. The two arms do not separate. Across three seeds, the paired difference between mined and random curricula is $-0.0045 \pm 0.0119$ AUPRC, $-0.0017 \pm 0.0314$ MCC and $-12 \pm 139$ false alarms, all spanning zero; the only difference clearing its interval is AUROC, at $-0.0013 \pm 0.0007$, and it \emph{favours the random control} at all three seeds. Neither curriculum improves on Stage 1 ($+0.0024 \pm 0.0253$ mined, $+0.0069 \pm 0.0182$ random).

At this curriculum size the intelligence of the mining step is not measurable, and the curriculum's effect comes from exposure balancing, which random selection supplies equally well. A plausible mechanism is that mining operates on the training split, where the Stage 1 encoder already achieves $99.9\%$ AUROC: the snippets that remain difficult under geographic shift are, by construction, not the ones that are difficult on data the model has already fit.

The obvious objection is that this null result reflects a limitation of the adapter rather than the absence of a mining signal: at $12.58$M trainable parameters, the refiner may simply be unable to express what the curriculum encodes. We tested this by unfreezing the last four transformer blocks of the otherwise frozen backbone in \emph{both} arms, a five-fold increase to $62.97$M trainable parameters, while holding the curricula fixed so that capacity is the only thing that varies. The null does not move: the paired difference is $-0.0045 \pm 0.0061$ AUPRC unfrozen against $-0.0040 \pm 0.0117$ frozen, a difference of differences of $-0.0005 \pm 0.0080$. The added capacity buys nothing in either arm on its own either, worth $+0.0004 \pm 0.0090$ to the mined curriculum and $+0.0010 \pm 0.0010$ to the random one, which is the rank saturation of Section~\ref{subsec:rank} reached through a different lever.

\subsection{Contrastive Regularization Has Little Left to Enforce}
\label{subsec:supcon}
Stage 3 applies a supervised contrastive objective to Stage 2 representations. Across three seeds it changes little in either direction. Compared to the baseline runs, paired per-seed, the model achieves an AUPRC difference of $+0.0002 \pm 0.0096$ against the mined Stage 2 initialization, $-0.0043 \pm 0.0069$ against the random control, and $+0.0026 \pm 0.0250$ against Stage 1. Its MCC$^{*}$ of $0.643 \pm 0.015$ is nominally the highest in Table~\ref{tab:performance_comparison}, but it overlaps the $r{=}4$ arm.

We previously reported that this stage degrades the high-recall regime. Replication does not support that claim: the stage is inert rather than harmful. Against Stage 2, AUROC changed by only $-0.0033 \pm 0.0127$ and $\text{FPR}_{95}$ by $+0.0095 \pm 0.0260$. The original claim was prompted by a single low outlier of $0.9795$, the lowest of three seeds, against a replicated mean of $0.9856$.

One possible explanation lies in the training loss itself. The loss floor $\mathbb{E}[\log |P(i)|]$ is $4.28$ at our batch size and target fraction, while a structureless embedding scores $5.34$ under the same projection. On the encoders it actually receives, Stage 3 begins at a loss between $4.40$ and $4.62$. That is $75$ to $91\%$ of the distance from a structureless state to a class-collapsed one, and the loss barely changes thereafter, so the preceding supervised stages have already imposed most of the geometry the contrastive term exists to enforce. Because the random control begins at a similar value ($4.55$), this geometry is a property of supervised adaptation itself rather than of the mined curriculum.

\subsection{Operating Point, Calibration and Threshold Transfer}
\label{subsec:operating_point}
The $85\%$ recall anchor is a reporting convention rather than an operational requirement, so Fig.~\ref{fig:pr_curves} and Fig.~\ref{fig:nfa_vs_recall} give the full trade-off. Fig.~\ref{fig:nfa_vs_recall} uses a logarithmic ordinate because the unadapted control's false-alarm load runs an order of magnitude above every adapted arm. The separation gained from domain adaptation holds across the entire recall range, while the individual adapted arms remain visually indistinguishable over most of the curve, the same conclusion the intervals of Table~\ref{tab:performance_comparison} reach. The false-alarm load rises sharply beyond about $0.9$ recall, so the marginal cost of the last few points of recall is steep and worth quoting in any operational requirement.

The later stages contribute one benefit invisible to the ranking metrics: expected calibration error falls from $0.0116$ at Stage 1 to $0.0030$ after Stage 2 and $0.0036$ after Stage 3, and the random control does \emph{not} reproduce this ($0.0152$), making calibration the one axis on which mining separates from random selection. Expected Calibration Error (ECE) is computed over $15$ equal-width bins on the confidence $\max(p, 1{-}p)$, with no temperature scaling or isotonic regression at any stage, so these are properties of the trained classifier and not of a post-hoc correction.

Threshold transfer is inexpensive in aggregate: the validation-tuned choice costs $1$ to $20$ false alarms more than an oracle threshold at the same recall, under $4\%$ of the total in every arm. Per platform it is not. Decomposing the blind split by sensor serial across three platforms, carrying $244$, $135$ and $52$ of the $431$ targets, the adapted model's recall spans $0.692$ to $0.906$ about a global $0.828$, so a deployment requiring $85\%$ recall on \emph{every} platform would not obtain it from an operating point whose aggregate satisfies the requirement. The spread is not a property of our three-stage framework: the frozen control ranges $0.763$ to $0.951$ about $0.886$ and TinyViT $0.635$ to $0.873$ about $0.773$, and the easiest platform is the same under all three architectures, which points at differences between the underlying surveys rather than at any model. Per-survey calibration is therefore worth its cost where a recall floor is contractual. We claim the spread and not a ranking: the ordering of the two weaker platforms reverses between the adapted and frozen arms, and the weakest carries only $52$ targets.

\subsection{Limitations of the Protocol}
\label{sec:limitations}
Two properties of the protocol above qualify the comparisons we have drawn. Neither changes a reported number, and we state both because a reader cannot recover either from the results table.

\textbf{The self-supervised arm sees more imagery than the arms it is compared against.} Contrastive pretraining is label-free, so we follow standard practice and fit the CC-SSL encoder over all 148 mission files, validation and test included, whereas every other arm sees the training missions alone. No annotation from the blind missions reaches any model, but that encoder has seen the unlabeled \emph{imagery}, so its row is transductive and not strictly like-for-like. A second asymmetry runs opposite, since it is scored through a linear probe on frozen features while the supervised arms are optimized end-to-end. The two are indistinguishable in AUPRC, and both asymmetries bear on that comparison in opposite directions.

\textbf{The validation split is used for three successive selection decisions.} Adapter rank is chosen on validation AUPRC, and Stages 2 and 3 each select their checkpoint by early stopping on the same split. Each is individually leakage-free with respect to the test missions, but they are not independent, they rest on only 362 validation targets, and the resulting optimism is not captured by the test metrics. The magnitude is not negligible: under a validation-tuned threshold, test recall drifted from $0.773$ to $0.926$ across arms, indicating substantial shift between validation and test missions before any selection is stacked on top. Reading our headline metrics at a recall-matched operating point insulates the false-alarm counts from this bias, but it does not protect the choice of rank or checkpoint. For replications, carve a dedicated selection partition out of the training missions and reserve the validation split solely for establishing the final operating point; rotating geographic folds between stages should be avoided, since doing so would confound stage effects with geographic variance.

\deleted{\added{Attempting to reproduce the predecessor contrastive SAS pipeline~\cite{Team_SAS_ViT_Unpub}, we identified three specification ambiguities: missing positive terms in the InfoNCE denominator, single-view negative pooling, and literal variance normalization ($x/\mathrm{var}(x)$). The conventional canonical configuration beats the literal as-typeset reading by a mean $+0.158$ AUPRC. Resolving ambiguities individually (Table~\ref{tab:repro_ablation}) reveals that loss formulation changes fall within seed noise, whereas \textbf{correcting data normalization accounts for 95\% of the recovery}. Literal variance division expands input range to roughly 2–69, disrupting pretrained expectations, whereas standard unit-variance normalization ($x/\sigma$) restores baseline performance.}}

\deleted{One further caveat belongs with these numbers. The two configurations were run on two different eight-GPU nodes, one variant per node, so reading is confounded with hardware. We checked this rather than assuming it away, repeating the as-typeset configuration bit for bit on the other node: at $r{=}4$ the two nodes agree to $0.010$ AUPRC, well inside the seed interval, and every cell of Table~\ref{tab:repro_ablation} and of the $+0.158$ mean is a low-rank cell. Under a \emph{full} fine-tune, however, the same check moved by $0.111$, four times the seed interval, which is why we exclude the full-fine-tune cells from the mean quoted above and report no reproduction claim that rests on them. Whether that instability is a property of the node or of full fine-tuning at this scale we cannot say from one seed.}

\section{Discussion}
\label{sec:discussion}
In this work we re-evaluated a three-stage parameter-efficient adaptation framework for SAS ATR, with every stage measured against its own matched control and every close comparison repeated across three seeds. The performance gains are dominated almost entirely by the initial domain-adaptation (LoRA) stage. That single step delivers a decisive $+0.38$ AUPRC improvement over a frozen backbone, and as little as $0.26\%$ of the model parameters is enough to obtain it.

Neither downstream stage separates from its control: hard negative mining is worth $-0.0045 \pm 0.0119$ AUPRC against a random curriculum of equal size, and supervised contrastive refinement $+0.0002 \pm 0.0096$ against the stage it refines. We report both null results in full, with the mechanisms established in Sections~\ref{subsec:hnm} and~\ref{subsec:supcon}; each follows from where the stage sits in the pipeline rather than from this dataset. The one axis on which mining does separate is calibration (Section~\ref{subsec:operating_point}).

Our replication also overturns an earlier reading of ours, that parameter efficiency requires backbone scale: no separation is resolved between low-rank adaptation and full fine-tuning at either scale we tested, and the rank curve is flat at both. Taken together, these results carry a concrete engineering implication. Effort is better spent optimizing the core adaptation step, and the operating point at which the system is evaluated, than on designing elaborate multi-stage training curricula, since the curriculum stacked on top of a well-adapted backbone did not repay its cost at this data scale.

Future work will extend the framework to Cluster-Level Contrastive Learning~\cite{ContrastiveClustering_2020} for unsupervised discovery of novel debris types, mine hard negatives from data the encoder has not already fit, and apply the contrastive objective earlier in training.

\bibliographystyle{./IEEEtran}
\bibliography{./IEEEabrv,./references}

\end{document}